\documentclass{clv3}
\usepackage{algorithmic}
\usepackage[ruled]{algorithm}
\usepackage{hyperref}
\usepackage{subfigure}
\usepackage{xcolor}
\definecolor{darkblue}{rgb}{0, 0, 0.5}
\hypersetup{colorlinks=true,citecolor=darkblue, linkcolor=darkblue, urlcolor=darkblue}

\begin{document}
	
	\title{Text Morphing}
	
	\author{Shaohan Huang$\dagger$ and Yu Wu$\dagger$}
	\affil{Microsoft Research}
	
	
	\author{Furu Wei~$\ddagger$\thanks{$\dagger$ Equal Contributions. $\ddagger$ Corresponding Author. \newline Natural Language Computing group, Microsoft Research Asia, Building 2, No. 5 Danling Street, Haidian District, Beijing, P.R. China 100080. E-mail: \{shaohanh, v-wuyu, fuwei, mingzhou@microsoft.com.\}}}
	\affil{Microsoft Research}
	
	\author{Ming Zhou}
	\affil{Microsoft Research}
	
	
	\maketitle
	
	\begin{abstract}
		In this paper, we introduce a novel natural language generation task, termed as \textbf{text morphing}, which targets at generating the intermediate sentences that are fluency and smooth with the two input sentences. We propose the Morphing Networks consisting of the editing  vector generation networks and the sentence editing networks which are trained jointly. Specifically, the editing vectors are generated with a recurrent neural networks model from the lexical gap between the source sentence and the target sentence. Then the sentence editing networks iteratively generate new sentences with the current editing vector and the sentence generated in the previous step. 
		We conduct experiments with 10 million text morphing sequences which are extracted from the Yelp review dataset. Experiment results show that the proposed method outperforms baselines on the text morphing task. We also discuss directions and opportunities for future research of text morphing.
	\end{abstract}
	
	\section{Introduction}
	Generating natural language sentences is a long-term vision and goal of natural language processing and has a broad range of real-life applications~\cite{nlg-survey-2018}. The mainstream methods for natural language generation usually generate a sentence from scratch. Recently, \citet{percyliang-edit} present the pioneering work on a new paradigm to generate sentences. Specifically, they propose a new generative model of sentences that first samples a prototype sentence from the training corpus and then edits it into a new sentence. Compared to traditional models that generate from scratch either left-to-right or by first sampling a latent sentence vector, the prototype-then-edit model improves perplexity on language modeling and generates higher quality outputs according to human evaluation.
	
	%
	
	\begin{table}[!h]
		\caption{An example of text morphing}	
		\label{example1} \small
		\centering
		\begin{tabular}{l}
			\hline
			\textbf{Source sentence:} The noodles and pork belly was  my favourite .  \\ \hline
			$s_1$: 	The pork belly was my favourite . \\ \hline
			$s_2$: 	The pork was very good . \\ \hline
			$s_3$: 	The staff was very good . \\ \hline
			$s_4$: 	The staff is very friendly . \\ \hline
			\textbf{Target sentence:} Love how friendly the staff is !\\ \hline

		\end{tabular}
		
	\end{table}
	
	We introduce a novel natural language generation task, termed as text morphing, which targets at generating the intermediate sentences that are fluency and smooth with the two input sentences. We show a concrete example of the text morphing task in Table \ref{example1} to elaborate on this task. Ideally, given a source sentence and a target sentence, our goal is to edit the source sentence step by step toward the target sentence where the source sentence is ``The noodles and pork belly was my favourite .'' and target sentence is ``Love how friendly the staff is !''. At the first step, we remove the term ``noodles'' from the source sentence, as it does not appear in the target sentence. After the deletion operation, $s_1$ is more closed to the target sentence in terms of the lexical similarity. The generated sentence $s_1$ is treated as the input of the second step. After two more editing operations, the editing process is terminated as a generated sentence $s_4$ is closed enough to the target sentence. Furthermore, we can find that the editing path is smooth because every editing operation only modifies a small part of the input sentence.
	
    To this end, we present an end-to-end neural networks model for generating morphing sentences between the source sentence and the target sentence. It consists of two parts, namely the editing vector generation networks and sentence editing networks. We design the editing vector generation networks to generate editing vectors with a recurrent neural networks model from the lexical gap between the source sentence and the target sentence. Then the sentence editing networks generate new sentences with the current editing vector and the sentence generated in the previous step. The two models are jointly trained and optimized.
	
	Text morphing provides a new direction to generate sentences. Different from traditional sentence generation that generates sentences from scratch, and text editing that generates sentences from a prototype sentence~\cite{percyliang-edit}, text morphing generates sentences from two anchor sentences, namely the source sentence and the target sentence.
	We conduct experiments with 10 million text morphing sequences which are extracted from the Yelp data set \cite{yelp-corpus}, that consists of 30 million review sentences on Yelp \footnote{\url{https://www.yelp.com}}. Experiment results show the effectiveness of the models. We also discuss directions and opportunities for future research of text morphing.
	
	\section{Related Work}
	
	Our work is related to text editing. \citet{percyliang-edit} propose a new generative model that first samples a prototype sentence from the training corpus and then edits it into a new sentence. Experiments on Yelp review corpus~\cite{yelp-corpus} and the One Billion Word Language Model Benchmark~\cite{ChelbaMSGBK13} show that the prototype-then-edit model improves perplexity on language modeling and generates higher quality outputs according to human evaluation.
	\citet{david-delete-networks} propose a framework for computer-assisted text editing. It applies to translation post-editing and paraphrasing. A human editor modifies a sentence by marking tokens they would like the system to change, and the system then generates a new sentence which reformulates the initial sentence by avoiding marked words. They demonstrate the advantage of their approach to translation post-editing and paraphrasing.
	\citet{word-morphing} describes how to use word embeddings trained with word2vec~\cite{word2vec} and A* search algorithm to morph between words.
	
	This work is also related to the image morphing work which has been widely studied in the image processing community. Morphing between two images is a special effect in motion pictures and animations that changes (or morphs) one image or shape into another through a seamless transition. Most often it is used to depict one person turning into another through technological means or as part of a fantasy or surreal sequence. The readers are referred to~\namecite{Wolberg1998} for a survey on image morphing. Our work focuses on morphing between two sentences which is different from image morphing.
	
	\section{Problem Statement}
	Our goal is to learn a generative model for text morphing.  In this section, we formulate this task mathematically, in which the input, output, and requirements of the task are defined clearly.
    
    Suppose that we have a data set $\mathcal{D} = \{(X_{i,start} , X_{i,1}\ldots X_{i,end}  )\}_{i=0}^N$, where $(X_{i,start} , X_{i,1}\ldots X_{i,end}  )$ is a sentence sequence that represents an existing path of changing one sentence into another through a seamless transition. The sequence satisfies the conditions that $ \forall j \in [1,end], s(X_{i,j-1},X_{i,j}) < \epsilon $, $ s(X_{i,j-1},X_{i,start}) > s(X_{i,j},X_{i,start})$ ,  and $s(X_{i,j-1},X_{i,end}) < s(X_{i,j},X_{i,end})  $, where $s(\cdot,\cdot)$ is an arbitrary text similarity metric.  We wish the transition is smooth, so $\epsilon $ is introduced to control the  degree of each sentence change. In addition, the change should move $X_{i,j}$ toward the target sentence $X_{i,J_i}$, as well as away from source sentence $X_{i,J_0}$, thus the last two conditions are added. Furthermore, $\forall i, X_i$ is a readable sentence that does not suffer from grammatical error.
	
	With such dataset $\mathcal{D}$, our goal is to learn a model $g(S,T)$ that is capable of performing sentence morphing from an arbitrarily sentence $S$ to $T$. During this process, we require the generated morphing path satisfies above conditions.

	\section{Morphing Networks} \label{morph:model}
	
	\subsection{Model Overview}
\begin{figure*}[t]		
		\begin{center}
			\includegraphics[width=13cm]{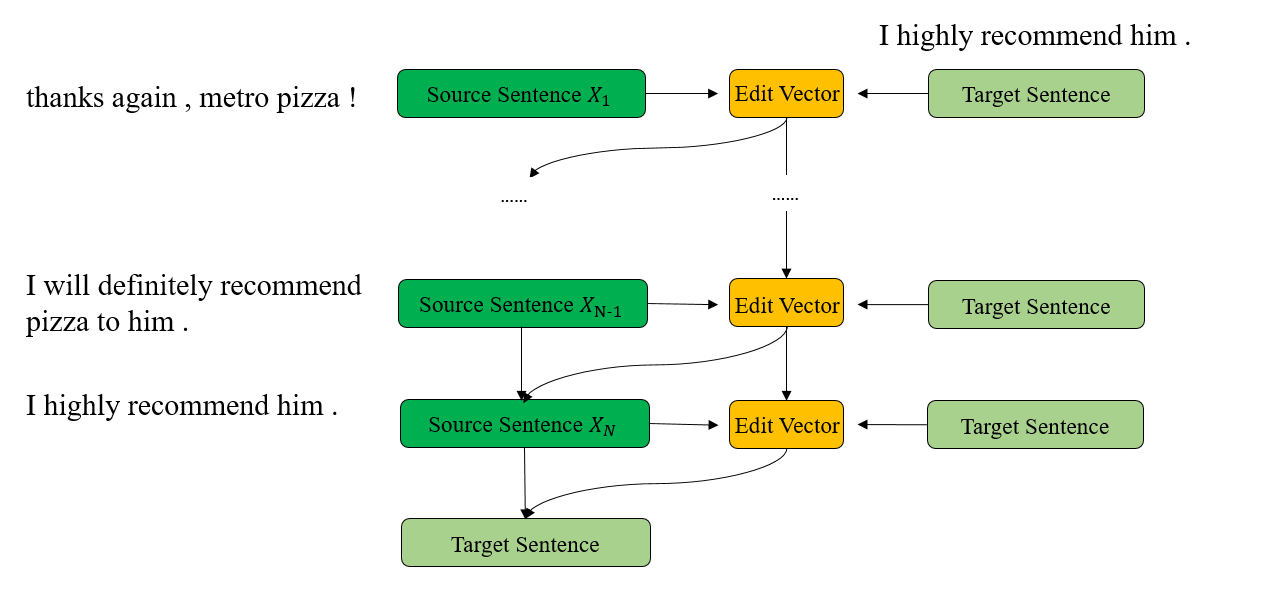}
		\end{center}
		\caption{The process of text morphing.}\label{fig:morphcase}
	\end{figure*}

	The target of text morphing is to generate the intermediate sentences that are fluency and smooth with the two input sentences. In practice, we design a morphing networks model that consists of editing vector generation and sentence editing. In particular, Figure \ref{fig:morphcase} depicts the iterative process of our model: 
	
	1. Editing vector generation: as the lexical gap between start sentence $X_{start}$ and end sentence $X_{end}$ is huge, we should determine which words will be edited at the current step, and then encode the information of these words into an editing vector. Two factors play a role in the editing vector generation. One is the lexical differences between $X_{start}$ and $X_{end}$, and the other is the editing vector of the last step. 
    
    2. Sentence editing: we further edit our source sentence $X_{i}$ with the editing vector and get a sentence $X_{i+1}$. After that, the next iteration begins with $X_{i+1}$ as the source sentence and edit it into another new sentence again. Through $N$-th editing, we obtain the fluent and smooth morphing path.

\begin{figure*}
		\begin{center}
			\includegraphics[width=13cm]{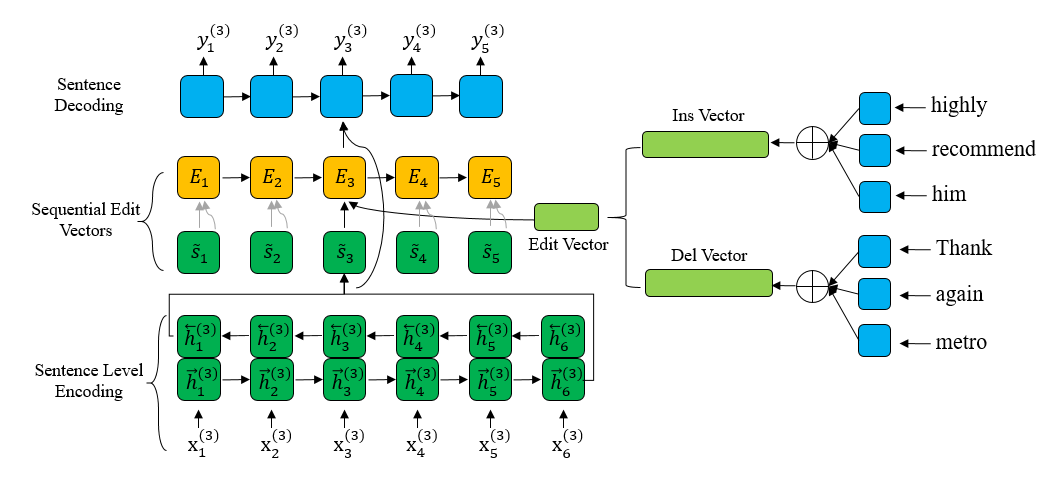}
		\end{center}
		\caption{Architecture of the morphing networks model.}
		\label{fig:morpharch}
	\end{figure*}

	Given a source sentence $X_{start} = (w_0,w_1 \ldots w_n )$ and a target sentence $X_{end} = (w'_0,w'_1 \ldots w'_n )$, our model generates a sentence sequence $(X_{start} , X_{1}\ldots X_{N} , X_{end})$ after editing $N$ step.
	The overview of our model is shown in Figure \ref{fig:morpharch}, we first compute the editing vector $E_i$ based on different words between $X_{start}$ and $X_{end}$ and the last editing vector $E_{i-1}$. We then build our sentence editing model on a left-to-right sequence-to-sequence model with attention, which integrates the edit vector into the decoder. In following, we will introduce the details of editing vector generation and sentence editing.
	
	\subsection{Editing Vector Generation}
	
	Given a source sentence $X_{start} = (w_0,w_1 \ldots w_n )$ and a target sentence $X_{end} = (w'_0,w'_1 \ldots w'_n )$, our model needs to edit $N$ steps to generate a morphing path $(X_{start} , X_{1}\ldots X_{N} , X_{end})$, which means that we will prepare $N$ editing vectors respectively. For each editing vector, there are two important factors to consider. One is the different words between source sentence $X_{start}$ and target sentence $X_{end}$ which provides the information which words will be edited at the current step. The other is the editing vector of the last step, which contains some information about which words have been edited. We leverage an RNN structure to generate the editing vector for each step, which can capture information about which words have been edited in the previous time steps.
	
	We first compute an insertion word set  $I = \{w | w\in X_{end} \wedge w\notin X_{start} \}$ where each element appears in $X_{end}$ but not in $X_{start}$,  and a deletion word set $D = \{w' | w'\in X_{start} \wedge w'\notin X_{end} \}$ where each element appears in $X_{start}$ but not in $X_{end}$. We look up an embedding table to transform words in $I$ and $D$ to dense vectors, forming an editing table $E = \{e_{i,0} \ldots e_{i,t}, e_{d,0} \ldots e_{d,t}\}$, where $e_{i,l}$ is the $l$-th insertion word embedding and $e_{d,l}$ is the $l$-th deletion word embedding. We use $E_i$ to denote the editing table of $X_i$ and $X_{end}$. 
	
	Given a sentence $X=(x_{1}, \ldots , x_{t})$, we learn its representation with a GRU based encoder which reads the input sentence $X$ into vectors like:
	\begin{equation} \label{decoderGRU}
	h_{t} = f_{\text{GRU}}(h_{t-1}, x_{t})
	\end{equation} where $x_{t}$ is the $t$-th word of $X$  and $h_{t}$  is a hidden state at time $t$. The hidden state of each word in sentence $X$ is $(h_{1}, \ldots, h_{t})$.
	
	After that, we apply the attention mechanism to generate a diff vector $d_i$. Specifically, the weight of $j$-th word in $D$ is computed by
	\begin{eqnarray} 
	&& \beta_{j} = \frac{exp(weight_{d,j})}{\sum_{j} exp(weight_{d,j})}, \\
	&& weight_{d,j} = \mathbf{v_{\beta}}^\top tanh(\mathbf{W_{\beta}}[e_{d,j} \oplus h_i]),
	\end{eqnarray}
	where $ \mathbf{v_{\beta}}$ and $\mathbf{W_{\beta}}$ are parameters, and $\beta_j$ is the weight of the $j$-th word in $D$. $h_i$  is the last hidden state of the encoder. The weight of $w_j$ in the insertion word set is obtained with the similar process, that is denoted as $\gamma_j$. Subsequently, we weighted average word embeddings to construct an insertion vector and a deletion vector separately, and then they are concatenated to form a diff vector $d_i$, which is formulated as
	\begin{equation}\label{att1}
	d_i =  \sum_{w \in I} \beta_w \Psi(w) \oplus \sum_{w' \in D} \gamma_{w'} \Psi(w'),
	\end{equation}	where $\oplus$ is a concatenation operation. Equation \ref{att1} plays an important role in our morphing model, because only a subset of the insertion/deletion word set will be used in each step. The larger weight is, the greater role of the word will play in the editing process. For instance, if a word in the deletion set is assigned with a large weight, the word is likely to be deleted in this step. 
	
	We employ a recurrent neural networks structure to compute the editing vector $z_{i}$, especially we use the gated recurrent unit (GRU) \cite{chung2014empirical} as the recurrent unit. When the prototype sentence is  $X_{i}$ and $h_i$ denotes the representation of prototype sentence $X_i$, the editing vector $z_{i}$ is defined as
	\begin{eqnarray}
	&& x'_i = h_{i} \oplus d_{i} \\
	&&z'_i = \sigma(\mathbf{W}_{z'}[x'_i, z_{i-1}]) \\
	&&r_i = \sigma(\mathbf{W}_r[x'_i,  z_{i-1}]) \\
	&&\widetilde{z}_i = \mathrm{tanh}(\mathbf{W}_h[x'_i, r_i \odot z_{i-1}]) \\
	&&z_i = (1-z'_i)\odot z_{i-1} + z'_i\odot \widetilde{z}_i
	\end{eqnarray}
	where $\oplus$ is a concatenation operation and $z_{i-1}$ is the last editing vector. $\mathbf{W}_{z'}$, $\mathbf{W}_r$ and $\mathbf{W}_h$ are parameters. The editing vector generation leverages the attention mechanism to determine which words will be encoded into an editing vector and the GRU structure to capture the information which words have been edited in the previous time steps.
	
	\subsection{Sentence Editing}
	
	We build our sentence editing model on a sequence-to-sequence with an attention mechanism model, which integrates the edit vector into the decoder.
	
	The decoder takes the encoder hidden state $(h_{1}, \ldots, h_{t})$ and edit vector $z_i$ as input and generate a new sentence by a GRU language model with attention. The hidden state of decoder is computed by 
	\begin{equation}
	h'_j = f_{\text{GRU}}(h'_{j-1}, y_{j-1} \oplus z_i)
	\end{equation} where $h'_{j-1}$ is the last step hidden state in decoder and we concatenate the word embedding of $(j\text{-}1)$-th word and editing vector $z_{i}$ as input.
	
	Then we compute a context vector $c_i$, which is a weighted sum of the input hidden states $(h_{1}, \ldots, h_{t})$ \cite{luong2015effective}:
	\begin{equation}\small
	c_j = \sum_{k=1}^t \alpha_{j, k} h_{k},
	\end{equation}
	where $\alpha_{j,k}$ is given by 
	\begin{eqnarray}\small \label{attention}
	&& \alpha_{j,k} = \frac{exp(e_{j,k})}{\sum_{l=1}^t exp(e_{j,l})}, \\
	&& e_{j,k} = \mathbf{v}^\top tanh(\mathbf{W_{\alpha}}[h_k\oplus h'_j]),
	\end{eqnarray}
	where $ \mathbf{v}$ and $\mathbf{W_{\alpha}}$ are parameters. 
	The generative probability distribution is given by
	\begin{equation}
	s(y_{j} ) = softmax(\mathbf{W_p} [y_{j-1} \oplus h'_{j} \oplus c_j] +\mathbf{b_p}), 
	\end{equation}
	where $\mathbf{W_{p}}$ and $\mathbf{b_{p}}$ are two parameters. We append the edit vector to every input embedding of the decoder in Equation \ref{decoderGRU}, so the edit information can be utilized in the entire generation process. 
	
	We aim at maximizing the likelihood of the generated sentences for training data $\mathcal{D} = \{(X_{i,start} , X_{i,1}\ldots X_{i,t}, X_{i,end}  )\}_{i=0}^N$. We learn our model by minimizing the negative log-likelihood (NLL) and the loss is computed by:
	\begin{equation} 
	\mathcal{L} =  -\sum_{i=0}^N \sum_{j=0}^t \text{log} p(X_{i, j+1}| X_{i, j}, z_j) 
	\end{equation} where $X_{i, 0}$ is $X_{i, start}$ and $X_{i, N+1}$ is $X_{i, end}$. 
	
	\section{Experiments} 
	\subsection{Dataset}
	We extract morphing sequences for training with the use of Yelp data set \cite{yelp-corpus}, that comprises of 30 million review sentences on Yelp \footnote{\url{https://www.yelp.com}}. Before extracting morphing sequences, we tokenize these sentences and replace named entities with their NER tags with spaCy \footnote{\url{https://honnibal.github.io/spaCy}}. Subsequently, we construct morphing dataset with the process shown in Algorithm \ref{alg:dataset}.

	We aim to could collect $N = 10$ million morphing instances. For each source sentence, we find at most $R = 10$ possible morphing sequences, since we wish the model is capable of learning different strategies for editing a sentence.  $\epsilon$ is set as $0.5$, that ensures the morphing path is smooth enough. Given a sentence, we use Minhash to search its similar sentences (Jaccard similarity is larger than $\epsilon =0.5$). An open source tool named as datasketch \footnote{\url{https://github.com/ekzhu/datasketch}} is employed to index sentences and find similar sentences. The hyperparameter, number of permutation, is chosen as $50$ which is a good trade-off between accuracy and efficiency.    $T_{min}$ and $ T_{max}$ are $4$ and $8$ respectively.
	\begin{algorithm}[t]
		\caption{Dataset Preparation for Text Morphing}	\label{alg:dataset}
		\begin{algorithmic}\small
			\STATE \textbf{input:} Sentence corpus $C$, valid range of a sentence sequence $[T_{min},T_{max}]$, morphing smoothness hyper-parameter $\epsilon$, empty dataset $D$, Instance number $N$, repeat number for a source sentence $R$.
			
			\WHILE{$l < N$}
			\STATE \textbf{Sample:} $ S \in C $
			\WHILE{ $j<R$}
			\STATE 	$X_{l,i} = S, i =0$
			\WHILE{ $i<T_{max}$}
			\STATE $ \mathbb{S} = \{X_{l,i} | X_{l,i} \in C \wedge J(X_{l,i},X_{l,i-1}) > \epsilon \wedge  J(X_{l,i},X_{0}) -  J(X_{l,i-1},X_{0}) > 0  \} $
			\IF {$\mathbb{S} $ is not empty}
			\STATE \textbf{Sample:} $X_i \in \mathbb{S}$, and append $X_i$ to $(X_{l,0} \ldots X_{l,i-1})$
			\STATE i = i + 1
			\ELSE
			\STATE break
			\ENDIF
			
			\ENDWHILE
			\IF {$i \in [T_{min},T_{max}]$}
			\STATE Add $(X_{l,0} \ldots X_{l,i})$ to $D$
			\STATE $l = l+1$
			\ENDIF
			\STATE j = j + 1
			\ENDWHILE
			\ENDWHILE
			\STATE \textbf{Output:}$D$
		\end{algorithmic}
	\end{algorithm}
	After removing duplications, we collect $9,956,038$ million morphing sequences, whose average sequence length is $6.67$. The average sentence length is 7.22, indicating that it is easier to find similar sentences for shorter text. 
	We randomly select $9,936,038$ for training, $10,000$ for validation and $10,000$ for test. We denote the test set as test set 1. 
	
	Apart from the validation set and testing set mentioned above, we randomly select $10,000$ sentences as source sentences and $10,000$ sentences as target sentences from the $30$ million Yelp dataset to construct another testing set. The testing set differs from the former one, since a morphing sequence cannot be obtained with a retrieval strategy. We denote the test set as test set 2. The dataset is available at \url{ https://1drv.ms/u/s!AmcFNgkl1JIngn4-tpg1yMYmh3bi}.
	
	\subsection{Evaluation Metrics}
	We evaluate this task in two aspects, fluency and smoothness. We train a 2-layer GRU based language model with 512 units on the 30 million Yelp dataset as a ruler of fluency. Given a morphing sequence, the metric reflects how fluent the sentences in the morphing sequence are, which is formulated as
    \begin{equation}
    fluency(X_{start}, X_1 \ldots X_{end}) = \sum_{ X_i \in (X_1 \ldots X_{end - 1})} f(X_i), 
    \end{equation} where $f(\cdot)$ denotes the negative log-likelihood probability of the sentence $X_i$. We average the fluency scores of morphing sequences as the final score. 
    
    An ideal morphing sequence is smooth, meaning two adjacent sentences ($ X_{i}$ and $X_{i+1}$) are similar on the lexicon. Jaccard distance is employed to calculate the smoothness of an editing operation. For a morphing sequence, we define two metrics, $Smoothness_{avg}$ and $Smoothness_{max}$, to indicate the smoothness of a morphing sequence as follows:
	
	\begin{equation}
Smoothness_{max}(X_{start}, X_1 \ldots X_{end}) = \max_{ X_i \in (X_1 \ldots X_{end - 1})} JaccardDis(X_{i-1},X_i).
	\end{equation}
		\begin{equation}
	Smoothness_{avg}(X_{start}, X_1 \ldots X_{end}) =  avg_{ X_i \in (X_1 \ldots X_{end - 1})} JaccardDis(X_{i-1},X_i).
	\end{equation}
	\subsection{Baselines}	
	We compare our method with following baselines:
	
	\textbf{Sentence VAE}: A variational autoencoder (VAE) \cite{kingma2013auto} allows us to generate sentences from a continuous space \cite{bowman2015generating}. Regarding to natural language morphing, we linear interpolate $N$ sentences between the source sentence $S$ and target sentence $T$, where $N$ is a hyper-parameter. Specifically, we first represent $S$ and $T$ with two latent vectors $z_1$ and $z_2$. $\forall t \in (0,N)$, the $t$-th latent vector is obtained by 
	\begin{equation}
	z_t = \frac{t}{N}  \cdot z_1 + (1- \frac{t}{N}) \cdot z_2. 
	\end{equation}
	Subsequently, a sentence is decoded from the vector with a GRU based decoder. In practice, we implement the baseline method with the open source code at  \url{https://github.com/timbmg/Sentence-VAE}, in which the KL cost annealing is applied to prevent  the decoder ignores $z$ and yields an undesirable stable equilibrium with the KL cost term at zero. SVAE is trained on the 30 million Yelp data. 
	
	\subsection{Implementation details}	
	We use PyTorch to implement our model.  The GRU hidden size is 512, word embedding size is 300, edit vector size is 256, and  attention vector size is 512. The vocabulary size is chosen as $30,000$.  We optimize the objective function using back-propagation and the parameters are updated by stochastic gradient descent with Adam algorithm \cite{kingma2014adam}.   The initial learning rate is $0.001$, and the parameters of Adam, $\beta_1$ and $\beta_2$ are $0.9$ and $0.999$ respectively. We employ early-stopping as a regularization strategy. Models are trained in mini-batches with a batch size of $128$. In the testing phase, we stop the editing process when $Jaccard(X_i, X_{end}) <= Jaccard(X_{i-1}, X_{end})$, or $Jaccard(X_i, X_{end}) >= 0.8$, or $i >= 10$. 
	\subsection{Experiment Results}	
	\begin{table}[!h]
		\caption{Evaluation results on test set 1. Lower fulency and smooth score is better.\label{eval1}}	
	
		\centering
		\begin{tabular}{l|c|c|c}
			\hline
			& fluency & $Smoothness_{max}$ & $Smoothness_{avg}$  \\ \hline
			SVAE              & 1.89 & 0.704 & 0.498\\ \hline
			Morphing Network & 2.56 & 0.372 & 0.235 \\ \hline

		\end{tabular}
		
	\end{table}

\begin{table}[!h]
	\caption{Evaluation results on test set 2. Lower fulency and smooth score is better.\label{eval2}}	
	
	\centering
	\begin{tabular}{l|c|c|c}
		\hline
		& fluency & $Smoothness_{max}$ & $Smoothness_{avg}$  \\ \hline
		SVAE              & 1.92 & 0.688 & 0.489\\ \hline
		Morphing Network & 3.00 & 0.399 & 0.232 \\ \hline

	\end{tabular}
	
\end{table}

The evaluation results are shown in Table \ref{eval1} and \ref{eval2}. Morphing Network is significantly better than SVAE method in terms of smooth metrics, indicating that the morphing network is able to transform a sentence to another with a series of small changes, whereas SVAE sometimes modifies a sentence massively in a morphing sequence. This is attributed to our model attends to a part of words at each step, so as to preserve other contents in the sentence. However, SVAE could generate more fluent sentences because SVAE is essentially a language model that pays more attention on fluency.  More experimental results of SVAE with different linear interpolate number N can be found in Table~\ref{appendix-svae} in Appendix. As the average number of inserted sentences is 3.6 in Morphing Network model, we report evaluation results of SVAE when N equals 4 in Table \ref{eval1} and \ref{eval2}.

	\subsection{Case Study}
\begin{figure*}[!h]
		\begin{center}
			\subfigure[Attention heat map of ($X_{start}, X_{end}$)\label{fig:att1}]	{
				\includegraphics[width=13cm]{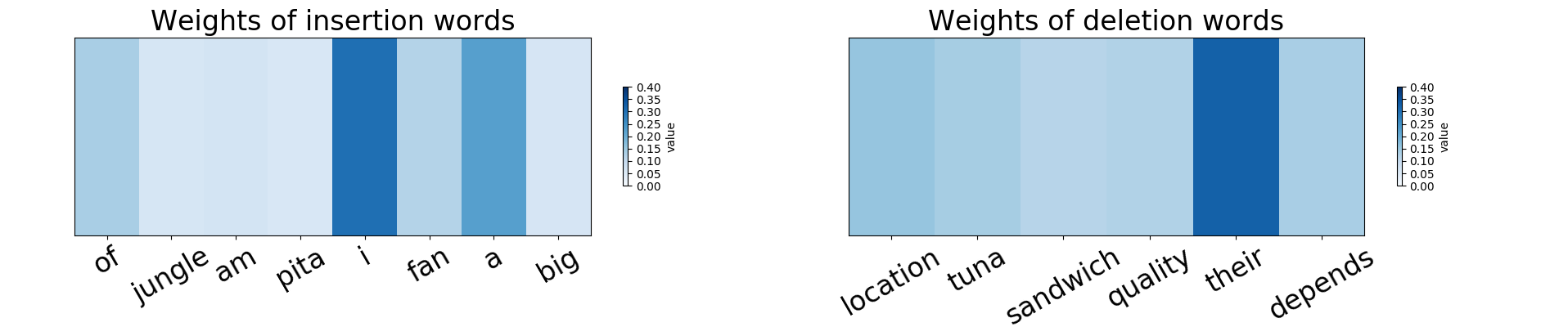}
			}
			\subfigure[Attention heat map of ($X_{1}, X_{end}$)	\label{fig:att2}]{
				\includegraphics[width=13cm]{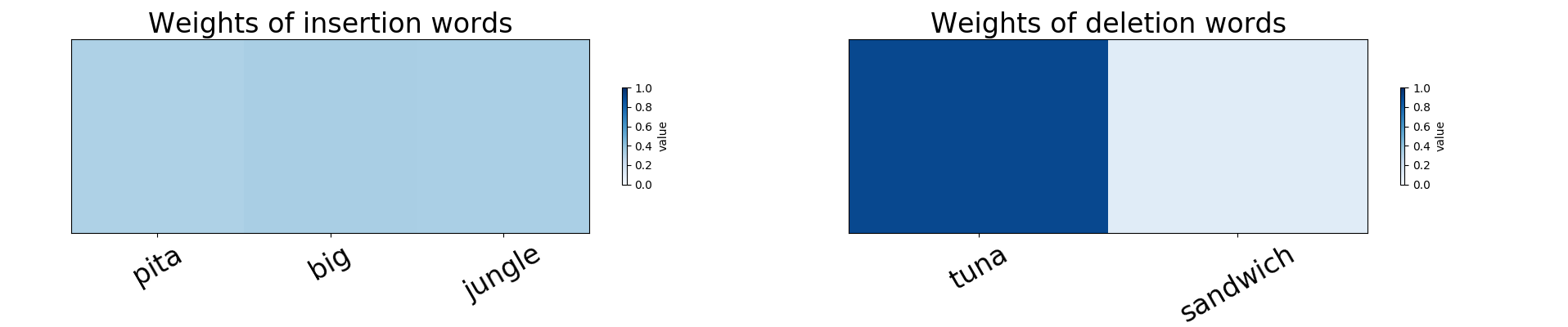}
			}
		\end{center}
		\caption{Edit attention heat map. 	\label{fig:att}}
		
	\end{figure*}

	\begin{table}[!h]
		\caption{Case Study}	
		\label{case} \small
		\centering
		\begin{tabular}{l}
			\hline
			\textbf{Source sentence:} 		their tuna sandwich quality depends the location .  \\ \hline
			$X_1$: 	i am a fan of the tuna sandwich. \\ \hline
			$X_2$: i am a big fan of the pita jungle . \\ \hline
			\textbf{Target sentence:} 	i am a big fan of the pita jungle .\\ \hline
			
			\textbf{Source sentence:} i opted for the wagyu filet .  \\ \hline
			$X_1$: 	i opted for the filet mignon . \\ \hline
			$X_2$: 	i loved the filet mignon . \\ \hline
			$X_3$: 	my friend loved the filet mignon . \\ \hline
			\textbf{Target sentence:} my friend loved the gluten free crust .\\ \hline

			\textbf{Source sentence:} 		the hot dishes were served piping hot .  \\ \hline
			$X_1$: 		the hot dishes were served hot \\ \hline
			$X_2$: the hot and hot dogs were hot and delicious . \\ \hline
			$X_3$: 	the hot dogs were hot and delicious . \\ \hline
			$X_4$: the hot dogs were hot and delicious , the service was great . \\ \hline
			$X_5$:the hot dogs were ok , the service was great . \\ \hline
			\textbf{Target sentence:} 	the service was great , and the food was ok .\\ \hline

		\end{tabular}
		
	\end{table}
	Table \ref{case} shows some text morphing examples given by our model. Our model is capable of transferring a sentence into another through a sequence of plausible sentences. In addition, our model can dynamically control the length of a morphing sequence rather than setting a hyper-parameter like SVAE. The three examples finish the text morphing by different times of text editing. The first example completes morphing with two revisions, and its editing attention heap map is depicted in Figure \ref{fig:att}.
	When we regard the source sentence as an input, ``I'', ``a'', ``of'' and ``fan'' get large weights in the insertion word attention, and ``their'', ``location'' and ``depends'' are top three in the deletion word set. Consequently, the source sentence is transformed into $X_1$, where words with large weights are inserted/deleted from the source sentence. Subsequently, the insertion word set and deletion word set are updated according to the differences between $X_1$ and the target sentence. The attention heat map of the second round editing is shown in Figure \ref{fig:att2}. We can find that the weights of inserted words are averaged, so all of these words are inserted into $X_1$. The weight of ``tuna'' dominates the deletion attention distribution, but ``sandwich'' and "quality'' are deleted from $X_1$ as well. This is mainly because that the decoder language model is forced to insert the phrase, ``pita jungle'', so the word, ``sandwich'', has to be deleted together with ``tuna'' to guarantee the fluency of the generated sentence.
	
	\section{Conclusion}
	
    In this paper, we introduce the \textbf{Text Morphing} task for natural language generation. It aims to generate the intermediate sentences that are fluency and smooth with the two input sentences. We present the Morphing Networks that consist of two parts. The editing vector generation network uses a recurrent neural networks model to generate editing vectors from the lexical gap between the source sentence and the target sentence. Then the sentence editing networks iteratively generate new sentences with the current editing vector and the sentence generated in the previous step. Experiments results on 10 million morphing sequences from the Yelp review dataset illustrate the effectiveness of the proposed models. 
    
    The work presented in this paper can be advanced from different perspectives. 
    First, it is very interesting to use the idea of in AlphaGo~\cite{Silver_2016, silver2017mastering} in designing the Sequential Editing Networks and the Morphing Networks. We can learn the policy networks and value networks to guide and control the strategy of generating the editing vectors in the process of text morphing. Second, as for application, we are interested in using morphing networks to conduct quantitative evaluation and identification of literal creativity of writings, such as literary masterpiece, articles, news and so on. We can use the morphing probability of the sentences in the writings to sentences in existing literature as a metrics of literal creativity. Moreover, a more general and open question is that ``\textit{ are any two sentences reachable through morphing with trainable morphing models?}'' We leave them as the future work on text morphing.
	
	\section{Appendix}
	
	\begin{table}[!h] \label{appendix-svae}
	\caption{Sentence VAE evaluation results on different numbers of linear interpolate sentences}	
	\centering
	\scalebox{0.85}{
    \begin{tabular}{c|c|c|c|c|c|c}
    \hline
       & \multicolumn{3}{c|}{Test set 1} & \multicolumn{3}{c}{Test set 2} \\ \hline
    N  & fluency & $Smoothness_{max}$ & $Smoothness_{avg}$  & fluency & $Smoothness_{max}$ & $Smoothness_{avg}$ \\ \hline
    1  &    1.88      & 0.608  & 0.608 &     1.89     & 0.630  & 0.630 \\ \hline
    2  &    1.91      & 0.522  & 0.710 &   1.90       & 0.519  & 0.713 \\ \hline
    3  &      1.90    & 0.650  & 0.581 &      1.89    & 0.638  & 0.579 \\ \hline
    4  &    1.89      & 0.704  & 0.498 &     1.92     & 0.688  & 0.489 \\ \hline
    5  &     1.90     & 0.717  & 0.425 &    1.89      & 0.719  & 0.427 \\ \hline
    6  &      1.89    & 0.760  & 0.382 &  1.90        & 0.729  & 0.379 \\ \hline
    7  &     1.91     & 0.753  & 0.336 &    1.91      & 0.774  & 0.344 \\ \hline
    8  &     1.91     & 0.764  & 0.307 &   1.90       & 0.762  & 0.307 \\ \hline
    9  &    1.89      & 0.764  & 0.276 &       1.89   & 0.778  & 0.287 \\ \hline
    10 &    1.91      & 0.785  & 0.257 &      1.89    & 0.779  & 0.258 \\ \hline
    \end{tabular}
    }
    \end{table}
	
	\starttwocolumn
	\bibliographystyle{compling}
	\bibliography{compling_style}

\begin{thebibliography}{15}
\expandafter\ifx\csname natexlab\endcsname\relax\def\natexlab#1{#1}\fi

\bibitem[{Bowman et~al.(2015)Bowman, Vilnis, Vinyals, Dai, Jozefowicz, and
  Bengio}]{bowman2015generating}
Bowman, Samuel~R, Luke Vilnis, Oriol Vinyals, Andrew~M Dai, Rafal Jozefowicz,
  and Samy Bengio. 2015.
\newblock Generating sentences from a continuous space.
\newblock \emph{arXiv preprint arXiv:1511.06349}.

\bibitem[{Chelba et~al.(2013)Chelba, Mikolov, Schuster, Ge, Brants, and
  Koehn}]{ChelbaMSGBK13}
Chelba, Ciprian, Tomas Mikolov, Mike Schuster, Qi~Ge, Thorsten Brants, and
  Phillipp Koehn. 2013.
\newblock One billion word benchmark for measuring progress in statistical
  language modeling.
\newblock \emph{CoRR}, abs/1312.3005.

\bibitem[{Chung et~al.(2014)Chung, Gulcehre, Cho, and
  Bengio}]{chung2014empirical}
Chung, Junyoung, Caglar Gulcehre, KyungHyun Cho, and Yoshua Bengio. 2014.
\newblock Empirical evaluation of gated recurrent neural networks on sequence
  modeling.
\newblock \emph{NIPS 2014 Deep Learning and Representation Learning Workshop}.

\bibitem[{Gatt and Krahmer(2018)}]{nlg-survey-2018}
Gatt, Albert and Emiel Krahmer. 2018.
\newblock Survey of the state of the art in natural language generation: Core
  tasks, applications and evaluation.
\newblock \emph{J. Artif. Intell. Res.}, 61:65--170.

\bibitem[{Grangier and Auli(2018)}]{david-delete-networks}
Grangier, David and Michael Auli. 2018.
\newblock Quickedit: Editing text {\&} translations via simple delete actions.
\newblock \emph{NAACL}.

\bibitem[{Guu et~al.(2018)Guu, Hashimoto, Oren, and Liang}]{percyliang-edit}
Guu, K., T.~B. Hashimoto, Y.~Oren, and P.~Liang. 2018.
\newblock Generating sentences by editing prototypes.
\newblock \emph{Transactions of the Association for Computational Linguistics
  (TACL)}.

\bibitem[{Kingma and Ba(2014)}]{kingma2014adam}
Kingma, Diederik~P and Jimmy Ba. 2014.
\newblock Adam: A method for stochastic optimization.
\newblock \emph{arXiv preprint arXiv:1412.6980}.

\bibitem[{Kingma and Welling(2013)}]{kingma2013auto}
Kingma, Diederik~P and Max Welling. 2013.
\newblock Auto-encoding variational bayes.
\newblock \emph{arXiv preprint arXiv:1312.6114}.

\bibitem[{Luong, Pham, and Manning(2015)}]{luong2015effective}
Luong, Minh-Thang, Hieu Pham, and Christopher~D Manning. 2015.
\newblock Effective approaches to attention-based neural machine translation.
\newblock \emph{arXiv preprint arXiv:1508.04025}.

\bibitem[{Mikolov et~al.(2013)Mikolov, Sutskever, Chen, Corrado, and
  Dean}]{word2vec}
Mikolov, Tomas, Ilya Sutskever, Kai Chen, Greg~S Corrado, and Jeff Dean. 2013.
\newblock Distributed representations of words and phrases and their
  compositionality.
\newblock In C.~J.~C. Burges, L.~Bottou, M.~Welling, Z.~Ghahramani, and K.~Q.
  Weinberger, editors, \emph{Advances in Neural Information Processing Systems
  26}. Curran Associates, Inc., pages 3111--3119.

\bibitem[{Silver et~al.(2016)Silver, Huang, Maddison, Guez, Sifre, van~den
  Driessche, Schrittwieser, Antonoglou, Panneershelvam, Lanctot, Dieleman,
  Grewe, Nham, Kalchbrenner, Sutskever, Lillicrap, Leach, Kavukcuoglu, Graepel,
  and Hassabis}]{Silver_2016}
Silver, David, Aja Huang, Chris~J. Maddison, Arthur Guez, Laurent Sifre, George
  van~den Driessche, Julian Schrittwieser, Ioannis Antonoglou, Veda
  Panneershelvam, Marc Lanctot, Sander Dieleman, Dominik Grewe, John Nham, Nal
  Kalchbrenner, Ilya Sutskever, Timothy Lillicrap, Madeleine Leach, Koray
  Kavukcuoglu, Thore Graepel, and Demis Hassabis. 2016.
\newblock Mastering the game of {Go} with deep neural networks and tree search.
\newblock \emph{Nature}, 529(7587):484--489.

\bibitem[{Silver et~al.(2017)Silver, Schrittwieser, Simonyan, Antonoglou,
  Huang, Guez, Hubert, Baker, Lai, Bolton, Chen, Lillicrap, Hui, Sifre, van~den
  Driessche, Graepel, and Hassabis}]{silver2017mastering}
Silver, David, Julian Schrittwieser, Karen Simonyan, Ioannis Antonoglou, Aja
  Huang, Arthur Guez, Thomas Hubert, Lucas Baker, Matthew Lai, Adrian Bolton,
  Yutian Chen, Timothy Lillicrap, Fan Hui, Laurent Sifre, George van~den
  Driessche, Thore Graepel, and Demis Hassabis. 2017.
\newblock Mastering the game of go without human knowledge.
\newblock \emph{Nature}, 550:354--.

\bibitem[{Wolberg(1998)}]{Wolberg1998}
Wolberg, George. 1998.
\newblock Image morphing: a survey.
\newblock \emph{The Visual Computer}, 14(8):360--372.

\bibitem[{Yelp(2017)}]{yelp-corpus}
Yelp. 2017.
\newblock Yelp dataset challenge.

\bibitem[{Zeldes(2018)}]{word-morphing}
Zeldes, Yoel. 2018.
\newblock Word morphing.

\end{thebibliography}
	
\end{document}